\documentclass{article}

    \PassOptionsToPackage{numbers, compress}{natbib}

    \usepackage[preprint]{neurips_2025}

\usepackage{amsmath}
\usepackage[utf8]{inputenc} 
\usepackage[T1]{fontenc}    
\usepackage[colorlinks=true,linkcolor=red,citecolor=green]{hyperref}
\usepackage{hyperref}       
\usepackage{url}            
\usepackage{algorithm}
\usepackage{algpseudocode}
\usepackage{booktabs}       
\usepackage{amsfonts}       
\usepackage{nicefrac}       
\usepackage{microtype}      
\usepackage{graphicx}
\usepackage{subfig}
\usepackage{float}
\usepackage{makecell}
\usepackage{multirow}
\usepackage{bbding}
\usepackage[table]{xcolor}
\usepackage{caption}
\usepackage{natbib}
\usepackage{stmaryrd}
\setcitestyle{numbers,square,citesep={,}}
\usepackage{enumitem}

\usepackage{marvosym}

\newcommand{\eg}{\textit{e.g.}}
\newcommand{\ie}{\textit{i.e.}}

\newcommand{\relu}{\text{ReLU}}
\newcommand{\prelu}{\text{PReLU}}
\newcommand{\img}[1]{{\mathbf #1}}
\newcommand{\conc}[2]{{\left\llbracket#1, #2\right\rrbracket}}
\newcommand{\vct}[1]{{\mathbf #1}}
\newcommand{\loss}[1]{\mathcal{L}_\text{#1}}
\definecolor{c2}{HTML}{FBD9BD}
\definecolor{c3}{HTML}{fe793d}
\definecolor{c4}{HTML}{eedeb0}
\definecolor{rouse}{rgb}{0.981,0.961,0.941}

\newcommand{\parhead}[1]{%
    \vspace{-2mm}\paragraph{#1}
}

\title{Uncertainty-Masked Bernoulli Diffusion for Camouflaged Object Detection Refinement}

\author{%
  Yuqi Shen$^{1,*}$
 \,,
  Fengyang Xiao$^{2,*}$\,,
  Sujie Hu$^{1}$\,,
  Youwei Pang$^{3}$\,,\\
  \textbf{Yifan Pu}$^{1}$\,,
  \textbf{Chengyu Fang}$^{1}$\,,
  \textbf{Xiu Li}$^{1}$\,,
  \textbf{Chunming He}$^{2}$\,,\\
  $^1$Tsinghua University, ~$^2$Duke University, ~$^3$Dalian University of Technology\\
}
\begin{document}
\maketitle
\begin{abstract}
Camouflaged Object Detection (COD) presents inherent challenges due to the subtle visual differences between targets and their backgrounds. While existing methods have made notable progress, there remains significant potential for post-processing refinement that has yet to be fully explored.
To address this limitation, we propose the Uncertainty-Masked Bernoulli Diffusion (UMBD) model, the first generative refinement framework specifically designed for COD.
UMBD introduces an uncertainty-guided masking mechanism that selectively applies Bernoulli diffusion to residual regions with poor segmentation quality, enabling targeted refinement while preserving correctly segmented areas.
To support this process, we design the Hybrid Uncertainty Quantification Network (HUQNet), which employs a multi-branch architecture and fuses uncertainty from multiple sources to improve estimation accuracy. This enables adaptive guidance during the generative sampling process. The proposed UMBD framework can be seamlessly integrated with a wide range of existing Encoder-Decoder-based COD models, combining their discriminative capabilities with the generative advantages of diffusion-based refinement.
Extensive experiments across multiple COD benchmarks demonstrate consistent performance improvements, achieving average gains of 5.5\% in MAE and 3.2\% in $F^\omega_\beta$ with only modest computational overhead. Code will be released.
\end{abstract}

\section{Introduction}\label{sec:introduction}
Camouflaged Object Detection (COD) aims to segment targets that are seamlessly blended into their surroundings, an inherently challenging task due to the minimal visual distinctions and high intrinsic similarity between targets and backgrounds~\cite{fan2021concealed_survey0,zhao2025deep-codsurvey1}. COD has recently garnered increasing attention and rapid development for its various applications, including biodiversity monitoring~\cite{wang2024depth_agricultural}, medical image analysis~\cite{fan2020pranetpolyp_segmentation}, and industrial defect detection~\cite{kumar2008computerdefect_detection}.

Traditional methods~\cite{he2019Traditionalcod} relying on manually designed models with hand-crafted feature extractors often struggle to adapt effectively in intricate scenarios of COD. Fueled by the rapid advancements in deep learning, a multitude of representative methods, predominantly CNN-based~\cite{fan2020CamouflageSINet} and Transformer-based~\cite{yin2024camoformer}, have achieved significant advances. Most adopt discriminative Encoder-Decoder architectures with various elaborately designed modules and strategies employed to counteract camouflage, such as enhancing edge information~\cite{he2023camouflaged_feder}, performing iterative optimization~\cite{he2025run}, or exploiting cues from multi-modal information~\cite{fang2025integratingmmcod}. Despite the considerable success of this paradigm, these discriminative methods frequently suffer from over-smoothing results~\cite{he2022tacklingover-smoothing1,he2023mitigating-smoothing2} and imprecise boundaries~\cite{xiao2024survey-codsurvey2}. Recently, diffusion models~\cite{ho2020denoisingddpm,he2024diffusion_survey1,croitoru2023diffusion_survey2}, by explicitly modeling data distributions, have emerged as promising alternatives by performing segmentation from a generative perspective, potentially 
addressing the aforementioned limitations with outstanding performance~\cite{chen2024camodiffusion,sun2025conditional_camodiffusiontpami,yang2025uncertaintyUGDM,chen2023diffusionCODdiffcod}. 


\begin{figure}
  \centering
  \includegraphics[width=1.0\textwidth]{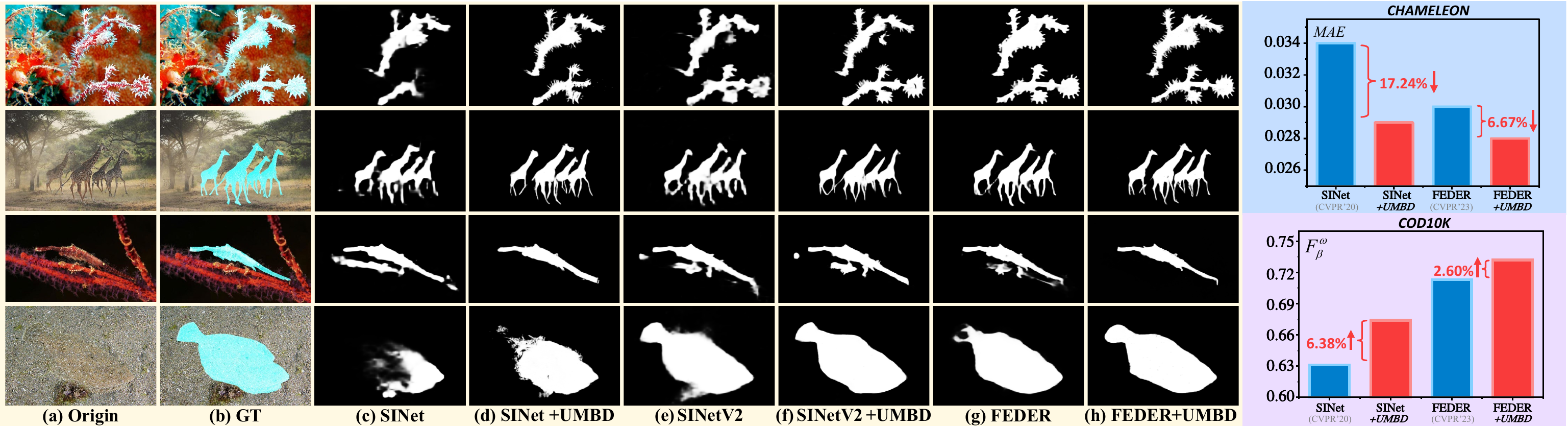}
  \small
  \vspace{-3mm} 
  \caption{Visual illustration of coarse masks from various COD methods and their corresponding refined outcomes by our method. (Row 1) Handles blurry edges; (Row 2) Recovers fine-grained details; (Rows 3-4) Reduces major false positives/negatives. The bar chart here quantitatively demonstrates the substantial improvements achieved. Please zoom in for a better view.}
  \label{fig:Visual_comparison_with_barchart}
  \vspace{-5mm} 
\end{figure}
 \vspace{-2.2mm} 
\noindent  
Although existing methods have pushed the boundaries of COD, the task's intrinsic difficulties mean there remains substantial room unexplored for improvement, particularly in delineation of extensive blurring edges, recovery of detailed structures, and mitigation of significant false positives/negatives (refer to Fig.~\ref{fig:Visual_comparison_with_barchart}). 
Existing refiners are mainly confined to conventional tasks (\ie, semantic segmentation~\cite{hao2020briefsemantic}) where coarse masks are generally well defined and largely complete. This limitation restricts their capability to minor refinements, consequently failing to meet the aforementioned critical refinement demands when adapted to COD tasks. More critically, their mundane designs exhibit insufficient discriminative power to handle the high foreground-background similarity, resulting in mismodifications of correct regions that even potentially deteriorating the initial segmentation quality.


To bridge this gap, we introduce the first widely compatible segmentation refiner, \underline{U}ncertainty-\underline{M}asked \underline{B}ernoulli \underline{D}iffusion (UMBD) model, specifically for COD, fully utilizing coarse masks and prior knowledge derived from existing Encoder-Decoder architecture models. We distinctively utilize Bernoulli Diffusion Models~\cite{sohl2015Bernoulli_theory1,AustinJHTB21Bernoulli_theory2}, capitalizing on their binary Bernoulli kernel's intrinsic alignment with segmentation tasks (inherently pixel-wise binary classification tasks) to reformulate refinement from a generative perspective~\cite{chen2024hidiff,wang2023segrefiner}. The iterative nature of diffusion models also aligns with coarse-to-fine refinement, enabling a synergistic integration of discriminative methods (providing prior knowledge) and generative methods (recovering misclassified regions). Furthermore, we propose a novel uncertainty-masked mechanism to customize the noise used in the Bernoulli diffusion process, employing a \underline{H}ybrid \underline{U}ncertainty \underline{Q}uantification \underline{N}etwork (HUQNet) that precisely estimates the uncertainty of coarse masks. This mechanism adaptively directs the model's focus towards uncertain areas, namely poorly segmented residual regions, facilitating more effective refinement while suppressing unnecessary confusion caused by global noise injection, particularly crucial given the inherent foreground-background similarity in COD tasks. The method demonstrates notable flexibility, enabling seamless integration with existing COD models for notable performance gains quantitatively and qualitatively (see Fig.~\ref{fig:Visual_comparison_with_barchart}). 
We further validate the generalizability of our refiner across broader concealed object scenarios through experiments on
polyp segmentation and transparent object detection tasks, also achieving substantial refinement gains as presented in Appx.~\ref{Extended Experiments}.

Overall, our main contributions are threefold:
\begin{enumerate}[leftmargin=*,itemsep=0.4em,topsep=0em,parsep=0em]
    \item We propose the first generative refiner, UMBD, explicitly designed for COD, which is compatible with the vast majority of existing COD methods and capable of leveraging their prior knowledge.
    \item We are the first to introduce Bernoulli diffusion to COD, characterized by a learnable uncertainty-masked mechanism to guide refinement by focused attention on challenging uncertain regions.
    \item Comprehensive experiments demonstrate our framework's strong compatibility with various COD methods, achieving average improvements of 5.5\% in MAE and 3.2\% in $F^\omega_\beta$ with consistent qualitative refinement. The superior performance on polyp/transparent object segmentation tasks further underscores the generalizability of our method across diverse segmentation scenarios.
    Leveraging the strengths of residual modeling, our method achieves notable refinement with only three sampling steps, substantially reducing the computational load in diffusion-based methods.
\end{enumerate}


\section{Related Work}\label{sec:Related-Work}

\parhead{Segmentation Refinement.}
The aim of segmentation refinement is to enhance the quality of masks in pre-existing models. Initial efforts develop model-specific refiners: PointRend \cite{pointrend} utilizes MLP to relabel low-confidence predictions. MaskTransfiner~\cite{transfiner} detects segmentation inconsistencies via an FCN module and refines labels through Transformer-based processing. Other works focus on universal refinement techniques~\cite{bpr, segfix, cascadepsp, polytransform,lin2025samrefiner}. SAMRefiner \cite{lin2025samrefiner} proposes a multi-prompt excavation strategy to extract diverse input prompts from coarse masks, which synergistically interact with SAM~\cite{kirillov2023SAM} for refinement. Notably, emerging approaches demonstrate Bernoulli Diffusion Models' effectiveness for segmentation refinement. SegRefiner~\cite{wang2023segrefiner} conducts preliminary explorations by directly applying Bernoulli diffusion models to segmentation refinement, and it solely utilizes coarse masks without fully leveraging prior knowledge from segmentation models. HiDiff~\cite{chen2024hidiff} introduces an optimized Bernoulli diffusion process for medical image segmentation. However, noise injection during the diffusion process confuses well segmented areas with uncertain regions under challenging scenarios (\eg, COD), ultimately resulting in incomplete denoised results. Notably, direct adaptation of existing refiners to COD tasks potentially yields paradoxically worse segmentation quality (shown in Sec.~\ref{sec:Comparison with Existing Refiners}), revealing a critical absence of an effective post-process segmentation refiner in the filed.

\parhead{Uncertainty Estimation.}
Uncertainty represents model ignorance about its prediction. Intuitively, uncertainty maps highlight regions where prior segmenters struggle, which need focused refinement attention. Theoretically, two primary types exist~\cite{zhang2023predictivePUENet}: epistemic uncertainty (model bias) and aleatoric uncertainty (data bias). Current uncertainty estimation methods primarily follow two paradigms: distribution-learning Bayesian neural networks~\cite{zhang2023predictivePUENet,yang2021uncertaintyUGTR} or simply fully-convolutional discriminators~\cite{yang2025uncertaintyUGDM,li2021uncertaintySCOD,zhang2022preynet,liu2022modeling}.
These methods typically integrate uncertainty awareness as auxiliary components or enhance outcome interpretability, thus requiring only coarse-grained estimations.

\begin{figure}
  \centering
  \includegraphics[width=1.0 \textwidth]{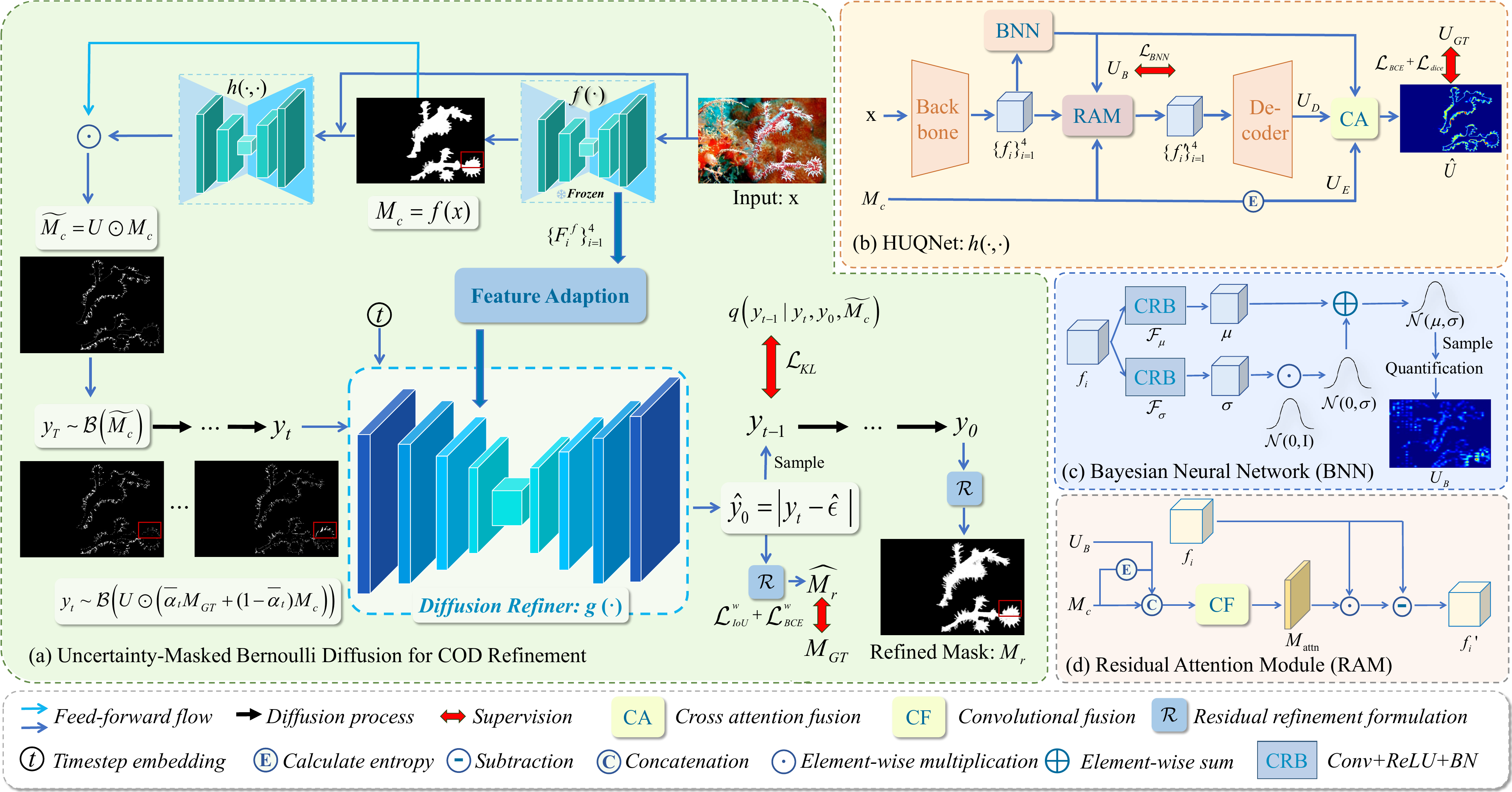}
  \small
  \vspace{-3mm}
  \caption{\textbf{UMBD Framework for COD Refinement}. The Bernoulli diffusion-based refiner leverages prior knowledge from pretrained models to perform uncertainty-masked interactive refinement, focusing on residual regions. Red boxes on the samples highlight regions where missing details in $M_C$ are progressively attended to and ultimately rectified in $M_r$ during the diffusion process. Panel (b) presents the proposed HUQNet with detailed modules in (c)-(d). Zoom in for a better view.}
  \label{fig:ModelArchitecture}
  \vspace{-3mm}
\end{figure}

\section{Methodology}\label{sec:Methodology}
In this section, we develop the first generative refiner UMBD for COD illustrated in Fig.~\ref{fig:ModelArchitecture}. Inspired by residual learning principles, our core insight is to focus the diffusion-based refiner on modeling the data distribution of poorly segmented residual regions, while preserving the correct regions from unnecessary modifications. This is realized via an explicit uncertainty-masked mechanism that spatially regulates latent variables conditioned on uncertainty maps, enabling region-specific noise injection rather than global diffusion, thereby achieving more targeted mask regeneration. This section proceeds with: Sec.~\ref{sec:preliminaries} introduces necessary preliminaries and notation. Sec.~\ref{sec:umbd_framework} details our UMBD refinement framework. Sec.~\ref{sec:huqnet} elaborates on the proposed HUQNet to estimate the uncertainty map needed during the inference stage.


\subsection{Preliminaries}\label{sec:preliminaries}

\parhead{Existing Segmenters and Residual Refinement Formulation.}
\label{sec:existing}
We utilize pretrained discriminative COD segmenters, denoted by $f(\cdot)$, which take an input image $\img{x} \in \mathbb{R}^{H \times W \times 3}$ and output a coarse mask $M_c=f(\img{x}) \in [0,1]^{H \times W \times 1}$. This initial prediction $M_c$ serves as the prior mask for our refinement process. The corresponding ground-truth mask is denoted as $M_{GT} \in \{0,1\}^{H \times W \times 1}$. Intuitively, let $U \in [0,1]^{H \times W \times 1}$ denote the uncertainty map highlighting challenging regions where $f(\cdot)$ produces inaccurate segmentation and
 its ground truth $U_{GT}$ can be defined as the pixel-wise L1 distance between $M_c$ and $M_{GT}$~\cite{yang2025uncertaintyUGDM,liu2022modeling}, namely $ U_{GT} = | M_c - M_{GT}|$. Conversely, $1-U$ corresponds to certain regions where $M_c$ aligns with $M_{GT}$. The ideal refined mask $M_r$ emerges from the residual refinement formulation in Eq.~\eqref{eq:refine_principle} when $U = U_{GT}$:
\begin{equation}
\label{eq:refine_principle}
    M_r = U \odot M_{GT} +  (1 - U) \odot  M_c,
\end{equation}
where $\odot$ denotes element-wise multiplication. Our diffusion process aims to learn the data distribution of the first residual term, while the proposed HUQNet $h(\cdot, \cdot)$ estimates $U$ in the formulation.

\parhead{Bernoulli Diffusion Models.}
Diffusion models consist of a forward and a reverse process.
The forward process $q(\img{y_{1:T}} | \img{y_0})$ uses a Markov chain to gradually convert the data distribution $\img{y_0} \sim q(\img{y_0})$ into a prior distribution (\eg, random Bernoulli noise), while the reverse process deploys a gradual denoising procedure $p_\theta (\img{y_{0:T}})$ that transforms random noise back into the original data distribution.
Compared to continuous diffusion models~\cite{T-DDPM,T-NCSN} based on Gaussian assumption, there is less research on Bernoulli diffusion models~\cite{sohl2015Bernoulli_theory1,AustinJHTB21Bernoulli_theory2}, where $y_T$ is typically defined to adhere to the Bernoulli distribution $\mathcal{B}(\img{y_T}|0.5)$. The forward process and reverse process are represented as:
\begin{gather}
    q(\img{y_{t}}|\img{y_{t-1}})      = \mathcal{B}(\img{y_t} | \img{y_{t-1}}(1-\beta_t)+0.5\beta_t), \quad
    p_\theta(\img{y_{t-1}}|\img{y_t}) = \mathcal{B}(\img{y_{t-1}} | \mu_\theta(\img{y_t}, t)),
\end{gather}
where $\beta_t \in (0,1)$ is a cosine noise schedule and $\mu_\theta(y_t, t)$ is a model predicting Bernoulli probability.
To enable the refiner to fully leverage $M_c$, the reverse process $p_{\theta}(\mathbf{y}_{t-1}|\mathbf{y}_{t}, f(\mathbf{x}))$ departs from classical diffusion models (initialized as $\mathcal{B}(\mathbf{y}_T|0.5)$) and instead starts denoising from $\mathbf{y}_{T} \sim \mathcal{B}(M_c)$~\cite{chen2024hidiff}.

\subsection{Uncertainty-Masked Bernoulli Diffusion for COD refinement}\label{sec:umbd_framework}

Based on the aforementioned preliminaries, our UMBD refiner utilizes a variant of UNet-based~\cite{ronneberger2015u} denoiser $g(\cdot)$ as the diffusion refiner to iteratively improve the coarse mask $M_c$ provided by a pretrained segmenter $f(\cdot)$ with frozen weights. 

\parhead{Uncertainty-Masked Forward Process.}
In the forward process, the uncertainty map is first applied to $M_{GT}$ and $M_c$ via $\img{{y}_0} =U \odot M_{GT}$ and $\widetilde{M}_c =U \odot M_C$. Then our methods constructs a forward diffusion process $q(\img{{y}_{1:T}}|\img{{y}_{0}}, \widetilde M_c)$, from $\img{{y}_0} \sim q(\img{{y}_0})$ to $\img{y}_T \sim \mathcal{B}(\widetilde{M}_c)$:
\begin{align}
\label{eq:forward}
    q\left(\img{y}_{1: T}|\img{y}_{0}, \widetilde{M}_c\right) &:= \prod\nolimits_{t=1}^{T}q\left(\img{y}_{t} |\img{y}_{t-1}, \widetilde{M}_c\right), \\ 
    \label{eqa:forward_tt-1}
    q\left(\img{y}_{t}|\img{y}_{t-1}, \widetilde{M}_c\right) &:= \mathcal{B}((1-\beta_{t}) \img{y}_{t-1}+\beta_{t}\widetilde{M}_c).
\end{align}
Using the notation $\alpha_{t}=1-\beta_{t}$ and $\bar{\alpha}_{t}= {\prod_{\tau=1}^{t}} {\alpha}_{\tau}$, we can sample $\img{y}_{t}$ at an arbitrary time step $t$ in a closed form:
\begin{align}
q\left(\img{y}_{t}|\img{y}_{0}, {M_c}, U\right) = \mathcal{B}\left(U \odot (\bar{\alpha}_{t} M_{GT} + (1-\bar{\alpha}_{t})M_c)\right).
    \label{eq:forward_arbitrary_step_sampling}
\end{align}
The Bernoulli sampling's mean parameter implements an uncertainty-modulated interpolation between $M_{GT}$ and the prior coarse mask $M_c$, where the interpolation weights evolve with timestep progression. Specifically, we can reparameterize $\img{y}_{t}$ in Eq.~\eqref{eq:forward_arbitrary_step_sampling} using Bernoulli-sampled uncertainty-aware customized noise. The reparameterization becomes:
\begin{align}\label{eq:reparameterization_forward_noise_addition}
 \mathbf{y}_t = \img{y_0} \oplus \mathbf{\epsilon}, \img{\epsilon} \sim \mathcal{B}((1-\bar{\alpha}_{t}) | \widetilde{M_c} - \img{{y}_0} |),  
\end{align}
where $\oplus$ denotes the logical operation of "exclusive OR". As derived from Eq.~\eqref{eq:reparameterization_forward_noise_addition}, our diffusion process constructs latent variables $\img{y}_t$ selectively injecting differentiated noise, correlated with uncertainty levels, into residual regions $U \odot M_{GT}$. This design explicitly encourages $g(\cdot)$ to reinforce its learning of refined knowledge for high-uncertainty regions. Without the uncertainty-masked mechanism, $\img{y}_t$ would degenerate into noise-blended certain regions, causing $g(\cdot)$'s misjudgment on refinement necessity, particularly problematic for COD tasks with high foreground-background similarity.
Notably, the forward process is only active during training (when $M_{GT}$ is accessible), thus $U$ is set to $U_{GT}$.
The Bernoulli posterior can be represented as:
\begin{align}
	q(\img{y}_{t-1}\!\mid\!\img{y}_{t}, \img{y}_{0}, \widetilde{M_c})\!=\!\mathcal{B}\Bigl(\phi_{\text{post}} \left( \img{y}_{t}, \img{y}_{0}, \widetilde{M_c} \right)\Bigr),\label{eq:posterior}
\end{align}
where $\phi_{\text{post}}(\img{y}_{t}, \img{y}_{0}, \widetilde{M_c}) = \|\{\alpha_{t}\conc{1-\img{y}_{t}}{\img{y}_{t}}+(1-\alpha_{t})|1-\img{y}_{t}-\widetilde{M_c}|\} \odot
    \{\bar{\alpha}_{t-1}\conc{1-\img{y}_{0}}{ \img{y}_{0}}+(1-\bar{\alpha}_{t-1})\conc{1-\widetilde{M_c}}{ \widetilde{M_c}}\}\|_1$~\cite{Hoogeboom2021ArgmaxFA}.
Here, $\|\cdot\|_1$ denotes the $\ell_1$ normalization along the channel dimension.

\parhead{Uncertainty-Masked Reverse Process.}
The reverse process \( p_{\theta}(\mathbf{y}_{0:T} \mid \widetilde{M}_c) \) is modeled as a Markov chain starting from \( \mathbf{y}_T \sim \mathcal{B}(\widetilde{M}_c) \), progressing through latent variables to capture the underlying data distribution:
\begin{align}
    &p_{\theta}(\img{y}_{0: T}|\widetilde{M_c}):=
    p(\img{y}_{T}|\widetilde{M_c})
     \prod\nolimits_{t=1}^{T} p_{\theta}(\img{y}_{t-1}|\img{y}_{t}, \widetilde{M_c}),\\
    &p_{\theta}(\img{y}_{t-1}|\img{y}_{t}, \widetilde{M_c}):=\mathcal{B}(\hat{\img{\mu}}(\img{y}_{t}, t, \widetilde{M_c})).
    \label{eqa:predicted_posterior}
\end{align}

The reverse process iteratively predicts the posterior mean using $g(\cdot)$ to estimate Bernoulli noise $\hat{\epsilon}(\img{y_t},t,\widetilde{M_c})$. Specifically, $\hat{\img{\mu}}(\img{y}_{t}, t, \widetilde{M_c})$ under the  $t$-th time step is reparameterized via Eq.~\eqref{eq:mu_hat}:
\begin{equation}\label{eq:mu_hat}
\begin{aligned}
    \hat{{\mu}}(\mathbf{y}_{t}, t,\widetilde{M_c}) =  \phi_{\text{post}}\left(\mathbf{y}_{t}, |\mathbf{y}_{t}-\hat{{\epsilon}}(\mathbf{y}_{t}, t, U \odot M_c )|, U \odot M_c  \right).
\end{aligned}
\end{equation}

The refined mask $M_r$ is derived from the final prediction $\hat{\img{y}}_0 = |\mathbf{y}_{t}-\hat{{\epsilon}}|$ in reverse process iteration through Eq.~\eqref{eq:M_r_estimation}:
\begin{equation}\label{eq:M_r_estimation}
    M_r = \hat{\img{y}}_0 + (1 - U) \odot M_c.
\end{equation}
Notably, during the training phase, we set the uncertainty map $U = U_{GT}$. While for the inference phase, we employ the estimated uncertainty map $U =\hat{U}$ obtained from our HUQNet $h(\cdot, \cdot)$ (detailed in Sec.~\ref{sec:huqnet}), where $\hat{U}=h(\img{x},M_c)$ . For implementation details of our UMBD refiner's training and inference sampling, please refer to the pseudo-code in Algorithms~\ref{alg:train},~\ref{alg:sample}.

\parhead{Feature Adaption.}
To further exploit the prior knowledge from pretrained $f(\cdot)$, the hierarchical features $\{F_i^f\}_{i=1}^4$ of the encoder in $f(\cdot)$ are fused with the features $\{F_i^g\}_{i=1}^4$ of $g(\cdot)$ via Eq.~\eqref{eq:resblocks}:
\begin{equation}\label{eq:resblocks}
F'^g_i = \mathcal{T}\Biggl( \prelu \circ\mathcal{T}\biggl( \Big( \psi^1_{3\times3}
\circ \mathcal{T}\big(\left\llbracket F^g_i, \mathcal{RB}(F^f_i,t)\right\rrbracket, t\big)  + \left\llbracket F^g_i, \mathcal{RB}(F^f_i,t) \right\rrbracket\Big)  
,t\biggr), t \Biggr),
\end{equation}
where $\circ$ denotes function composition, $\left\llbracket \cdot,\cdot\right\rrbracket$ denotes concatenation, $\psi_{k\times k}^s$ denotes a convolution operation with kernel size $k$ and stride $s$, $\mathcal{T}(\cdot,t)$ denotes time-conditioned bias modulation, and $\mathcal{RB}$ represents the ResBlock transformation detailed in Appx.~\ref{app:ResBlock-transformation}.

\parhead{Optimization and Inference.}
Following~\cite{chen2024camodiffusion,chen2024hidiff}, we optimize $g(\cdot)$ with $\loss{Diff}$, consisting of KL divergence, weighted intersection-over-union loss, and weighted binary cross entropy loss:
\begin{align}
     \loss{KL} &= \mathrm{KL}[q(\img{y}_{t-1} | \img{y}_{t}, \img{y}_{0}, \widetilde{M_c}) || p_{\theta}(\img{y}_{t-1} | \img{y}_{t}, \widetilde{M_c})],
\\
    \loss{Diff} &= \loss{KL} + \mathcal{L}^w_{\text{IoU}}(\hat{M}_r, M_{GT}) + \mathcal{L}^w_{\text{BCE}}(\hat{M}_r, M_{GT}),\label{eq:loss_UMBD}
\end{align}
where \( \hat{M}_r \) is obtained via Eq.~\eqref{eq:M_r_estimation}, with \( 
\hat{\img{y}}_0 = |\mathbf{y}_t - \hat{\epsilon}| \) and \( \hat{\epsilon} \) denoting the noise prediction from \( g(\cdot) \) at the current sampled training timestep as shown in Alg.~\ref{alg:train}.
As detailed in Alg.~\ref{alg:sample}, during inference we first generate a coarse mask and corresponding uncertainty estimation using $f(\cdot)$ and $h(\cdot, \cdot)$, respectively. Our UMBD then samples the initial latent variable $\img{y}_T$ from $\widetilde{M_c}$, followed by iterative refinement to obtain $M_r$. To accelerate sampling, we implement the DDIM strategy~\cite{T-DDIM} with $ \sigma_t = \frac{1 - \bar{\alpha}_t}{1 - \bar{\alpha}_{t-1}} $ to reduce reverse process stochasticity.

\algrenewcommand\algorithmicindent{0.5em}%
\begin{figure}[t]
  \renewcommand{\arraystretch}{0.5}
  \begin{minipage}[t]{0.495\textwidth}
  \begin{algorithm}[H]
        \caption{Training} \label{alg:train}
    \small
    \begin{algorithmic}[1]
      \Require image-mask pairs $\{ \mathcal{X}, \mathcal{Y} \}$
      \Repeat
          \State \textbf{Sample} $(\img{x},M_{GT})\sim\{ \mathcal{X}, \mathcal{Y} \}$
          \State \textbf{Calculate} $M_c=f(\img{x}), U_{GT} = | M_c - M_{GT}|$
          \Statex \qquad\qquad \ $\img{{y}_0} =U_{GT} \odot M_{GT}$, $\widetilde{M}_c =U_{GT} \odot M_C$
          
          \State \textbf{Sample} $t \sim \mathrm{Uniform}(\{1, \dotsc, T\})$
          \Statex \qquad\qquad $\img{\epsilon} \sim \mathcal{B}((1-\bar{\alpha}_{t}) | \widetilde{M_c} - \img{{y}_0} |)$
          \State \textbf{Calculate} $\mathbf{y}_t = \img{y_0} \oplus \mathbf{\epsilon}$
          \State \textbf{Estimate} $\hat{{\epsilon}}(\mathbf{y}_{t}, t, \widetilde{M_c})$
          \State \textbf{Calculate} $\hat{\img{y}}_0 = |\mathbf{y}_{t}-\hat{{\epsilon}}|$ 
          \Statex \qquad\qquad \ \, $\hat{M}_r = \hat{\img{y}}_0 + (1 - U_{GT}) \odot M_c$
          \Statex \qquad\qquad \ \ $q(\img{y}_{t-1}\!\mid\!\img{y}_{t}, \img{y}_{0}, \widetilde{M_c})$, $p_{\theta}(\img{y}_{t-1}|\img{y}_{t}, \widetilde{M_c})$
          \State Take gradient descent step on
          \Statex $\qquad {\nabla}_\theta \loss{Diff}$ while freezing $f(\cdot)$
      \Until{converged}
      \vspace{0.35em}
    \end{algorithmic}
  \end{algorithm}
  \end{minipage}
  \hfill
  \begin{minipage}[t]{0.495\textwidth}
  \begin{algorithm}[H]
      \caption{Sampling} \label{alg:sample}
    \small
    \begin{algorithmic}[1]
          \Require image $\img{x}$
      \State \textbf{Calculate} $M_c=f(\img{x})$
      \Statex \qquad\qquad \ \ $\hat{U}=h(\img{x}, M_C)$
      \Statex \qquad\qquad \  $\widetilde{M_c} = \hat{U} \odot M_c$
      \State \textbf{Sample} $\img{y}_T \sim \mathcal{B}(\widetilde{M_c})$
      \For{$t=T, \dotsc, 1$}
          \State Calculate $\hat{{\epsilon}}(\mathbf{y}_{t}, t, \widetilde{M_c})$
        \State When using DDPM’s sampling strategy:       
           \Statex \hspace*{1em} $\img{y}_{t-1} \sim \mathcal{B}(\hat{\img{\mu}})$, $\hat{{\mu}} =  \phi_{\text{post}}\left(\mathbf{y}_{t}, |\mathbf{y}_{t}-\hat{{\epsilon}}|,\widetilde{M_c} \right)$
          \State When using DDIM’s sampling strategy:       
           \Statex \hspace*{1em} $\vct{y}_{t-1}\!\sim\! \mathcal{B}(\sigma_{t}\vct{y}_{t}+(\bar{\alpha}_{t-1}\!-\!\sigma_{t}\bar{\alpha}_{t})|\vct{y}_{t}\!-\!\hat{\vct{\epsilon}}|+((1-\bar{\alpha}_{t-1})-(1-\bar{\alpha}_{t})\sigma_{t})\widetilde{M_c})$
      \EndFor
      \State \textbf{Calculate} $M_r = \hat{\img{y}}_0 + (1 - \hat{U}) \odot M_c$
      \State \textbf{return} $M_r$
    \end{algorithmic}
  \end{algorithm}
  \end{minipage}
\vspace{-3mm}
\end{figure}

\subsection{HUQNet for Uncertainty Estimation}\label{sec:huqnet}

We propose HUQNet $h(\cdot, \cdot)$ for uncertainty estimation, which outputs $\hat{U}= h(\img{x}, M_c)$ serving as the approximate surrogate for $U_{GT}$ required during the inference sampling stage. This hybrid model comprises a Bayesian Neural Network (BNN) branch and a discriminative branch, with a shared backbone. 
Leveraging a shared backbone and the fusion of estimations, our dual-branch design enables distribution modeling to perceive holistic patterns and discriminative methods to refine local details, with these mutually enhancing each other to achieve finer results and uncover previously neglected regions.
The pipeline and modules are illustrated in Fig.~\ref{fig:ModelArchitecture}(b,c,d).

\parhead{BNN Branch.}
Inspired by~\cite{yang2021uncertaintyUGTR,yang2025uncertaintyUGDM}, our proposed HUQNet first employs a ResNet50 backbone to extract hierarchical features $\{f_i\}_{i=1}^{4}$ from the input $\img{x} \in \mathbb{R}^{H\times W \times 3 }$. Following the practice in~\cite{yang2021uncertaintyUGTR}, the BNN branch of our hybrid model measures uncertainty scores $c$ for each pixel from a probabilistic representation perspective via Gaussian parameterization:
$c = \mu + \epsilon \cdot \sigma$ and $\epsilon \sim \mathcal{N}(0,1)$,
where \(\mu,\sigma\) denote learned mean and variance maps. As illustrated in Fig.~\ref{fig:ModelArchitecture}(c), the BNN branch processes backbone features \({f}_{3}\) through parallel \(\mu\)- and \(\sigma\)- feature projections:
$\mu = \mathcal{F}_\mu({f}_{3})$, 
$\sigma = \mathcal{F}_\sigma({f}_{3})$.
Uncertainty maps \({U}_B\) are derived from \(K\) Monte Carlo samples:
${U}_B = \text{Norm}\big(\text{Var}(\{c^{(k)}\}_{k=1}^K)\big)$,
where $\text{Norm}(\cdot)$ denotes the mean-max normalization and $\text{Var}(\cdot)$ is the variance calculation operation.

\parhead{Discriminative Branch.}
To empower the uncertainty quantification decoder to fully leverage existing information and effectively explore regions beyond current uncertainty estimations (\eg, $U_B$), we employ a Residual Attention Module (RAM, illustrated in Fig.~\ref{fig:ModelArchitecture}(d)) that adaptively modulates features delivered to the decoder through fusion of multi-source uncertainty estimations.
Considering the entropy map ($U_E$) of $M_c$ also serves as an effective uncertainty estimation, we concatenate the existing information $M_c$, $U_E$, and $U_B$ to form RAM's input $M_{\text{cat}} \in \mathbb{R}^{H\times W \times 3 }$. The convolutional fusion module generates an attention map $M_{\text{attn}}$ via a classical stacked conv-block design, detailed in Appx.~\ref{app:stacked_CF}.
After aligning its resolution with $\{f_i\}_{i=1}^{4}$ via bilinear interpolation, $M_{\text{attn}}$ is processed by a residual attention mechanism (Eq.~\eqref{eq:RAM}) for refined features $\{f'_i\}_{i=1}^{4}$.
\begin{equation}
    f'_i = f_i \odot \left(1 - M_{\text{attn}}\right).
    \label{eq:RAM}
\end{equation}

\parhead{Decoder.} 
Follow the design in~\cite{yang2025uncertaintyUGDM}, the multilevel features $\{f'_i\}_{i=1}^4$ modulated by RAM are progressively refined from level 4 to 1. Let $U_4 = f'_4$ denote the initial highest-level feature. The refined features $\{U_i\}_{i=1}^{3}$ are iteratively computed based on
$U_i = \text{DAB}\left(\psi_{3\times3}^1\left( \left\llbracket\text{TConv}(U_{i+1}), f'_i\right\rrbracket\right)\right)$,
where $\text{TConv}(\cdot)$ denote transposed convolution and $\text{DAB}(\cdot)$ is a composite block consisting of Dropout, Leaky ReLU activation, and Batch Normalization.
The final output of the discriminative branch $U_D$ is the lowest-level feature $U_1$.

\parhead{Efficient Cross-Attention Fusion.}
The efficient cross-attention module fuses the multi-source uncertainty maps (\ie, $U_E$, $U_B$, $U_D$) to obtain the final result $\hat{U}$ through window-based cross attention with reduced computational overhead.
%
%
%
The operation proceeds in several steps. First, window partitioning is applied to the input $U_D$ and $\left\llbracket U_E, U_B\right\rrbracket$ to obtain $\mathcal{W}_q = \Phi_{\text{win}}(U_D)$ and  $\mathcal{W}_{kv} = \Phi_{\text{win}}(\left\llbracket U_E, U_B\right\rrbracket)$, where $\Phi_{\text{win}}$ denotes $16\times16$ window partitioning. These windowed patches are then transformed via projections into query, key, and value matrices: $Q = \psi_{1\times1}^{1(q)}(\mathcal{W}_q)$, $K = \psi_{1\times1}^{1(k)}(\mathcal{W}_{kv})$, and $V = \psi_{1\times1}^{1(v)}(\mathcal{W}_{kv})$. Next, multi-head cross attention is computed as $\hat{V} = \text{Softmax}\left(\frac{QK^\top}{\sqrt{d_h}}\right)V$, with head dimension $d_h = 4$. After normalization via $\tilde{V} = \text{RMSNorm}(\hat{V}) + \hat{V}$, the final result $\hat{U}$ is obtained via window merging and a residual connection: $\hat{U} = \psi_{1\times1}^{1(out)}(\Phi_{\text{merge}}(\tilde{V})) + U_D$, where the residual formulation preserves results from the discriminative branch $U_D$ while integrating complementary information from $U_E$ and $U_B$.

\parhead{Optimization.}
Following~\cite{yang2021uncertaintyUGTR}, the training objective of BNN brach combines reconstruction (binary cross entropy loss) and distribution regularization (KL divergence):
\begin{equation}\label{eq: BNNloss}
\loss{BNN} = \mathcal{L}_{\text{BCE}}(c^{(k)}, M_{GT}) + \eta\cdot \mathcal{D}_{\text{KL}}(\mathcal{N}(\mu,\sigma)\|\mathcal{N}(0,1)),
\end{equation}
where \(\eta\) is empirically set to 0.1 to emphasize the model’s prediction, and \(c^{(k)}\) is a sample randomly drawn from the learned distribution.
Follow the practice of ~\cite{yang2025uncertaintyUGDM,liu2022modeling}, we optimize the discriminative branch with binary cross entropy loss and dice loss. Then the overall loss function of HUQNet is:
\begin{equation}\label{eq: HUQNetloss}
\loss{H} = \mathcal{L}_{\text{BCE}}(\hat{U}, U_{GT})+\mathcal{L}_{\text{dice}}(\hat{U}, U_{GT}) +\loss{BNN}.
\end{equation}



\section{Experiments}\label{sec:Experiments}

\subsection{Experimental Settings}\label{sec:experimental settings}

\parhead{Implementation Details.}\label{sec:Implementation_Details}
We implement UMBD in PyTorch with three RTX 4090 GPUs. Input resolutions follow the configurations of each integrated method with a batch size set to 36. We employ the AdamW optimizer~\cite{loshchilov2017decoupled} with polynomial decay scheduling.
Following~\cite{yang2021uncertaintyUGTR,chen2024hidiff}, initial learning rates are empirically set as denoiser ($1\!\times\!10^{-4}$), HUQNet backbone (ResNet50~\cite{he2016deepresnet50} pretrained on ImageNet~\cite{deng2009imagenet}, $1\!\times\!10^{-7}$), BNN branch ($1\!\times\!10^{-6}$), and the remaining modules in HUQNet ($1\!\times\!10^{-3}$). A cosine noise schedule~\cite{nichol2021improved} with $T=1000$ timesteps is applied for training.
For inference, we adopt DDIM sub-sequence sampling with $T=10$ steps, consistent with existing diffusion-based COD methods~\cite{yin2024camoformer,yang2025uncertaintyUGDM,chen2023diffusionCODdiffcod}. All results report averages over five runs with different random seeds.

\parhead{Training Strategy.}
To avoid overfitting during training $g(\cdot)$, its validation stage requires inference sampling using HUQNet's output $\hat{U}$. We therefore adopt a three-stage training strategy: (1) Pre-train HUQNet for 80 epochs until near convergence; (2) Train $g(\cdot)$ to convergence while maintaining frozen HUQNet parameters during validation phases to ensure stable sampling; (3) Fine-tune HUQNet to convergence with $g(\cdot)$ frozen, achieving optimal overall performance of the integrated system.

\parhead{Datasets and Metrics.}\label{Sec:Dataset}
We conduct experiments on COD tasks, adhering to the protocols established in~\cite{fan2020CamouflageSINet} and performing experiments on four datasets: \textit{CHAMELEON}~\cite{skurowski2018animal}, \textit{CAMO}~\cite{le2019anabranch}, \textit{COD10K}~\cite{fan2021concealed_survey0}, and \textit{NC4K}~\cite{lv2021simultaneously}.
For training, we use 1,000 images from \textit{CAMO} and 3,040 images from \textit{COD10K}. The remaining images from these two datasets, along with all images from the other datasets, constitute the test set.
To evaluate performance, we employ four widely-used metrics: mean absolute error ($M$), weighted F-measure ($F^\omega_\beta$)~\cite{margolin2014evaluate}, adaptive E-measure ($E_\phi$)~\cite{fan2021cognitive}, and structure measure ($S_\alpha$)~\cite{fan2017structure}. Superior performance is indicated by lower values of $M$ and higher values of $F^\omega_\beta$, $E_\phi$, and $S_\alpha$.

\subsection{Results and Comparative Evaluation}\label{sec:comparative_results}

\begin{table*}[!b]
    \setlength{\abovecaptionskip}{0cm}
    \setlength{\belowcaptionskip}{-0.2cm}
    \centering
    \caption{Quantitative performance gains of COD methods using diverse backbones with our proposed UMBD Refiner. Average improvement rates (\%) for each backbone group are marked in \textbf{bold}.}
    \label{table:Results_COD}
    \vspace{1mm}
    \resizebox{\textwidth}{!}{
    \setlength{\tabcolsep}{1.4mm}
    \begin{tabular}{l|cccc|cccc|cccc|cccc}
        \toprule
        \multicolumn{1}{c|}{\multirow{2}{*}{Methods}} 
        & \multicolumn{4}{c|}{\textit{CHAMELEON}} 
        & \multicolumn{4}{c|}{\textit{CAMO}} 
        & \multicolumn{4}{c|}{\textit{COD10K}} 
        & \multicolumn{4}{c}{\textit{NC4K}} \\ 
        \cline{2-17}
        & {\cellcolor{gray!40}$M$$\downarrow$} 
        & {\cellcolor{gray!40}$F^\omega_\beta$$\uparrow$} 
        & {\cellcolor{gray!40}$E_\phi$$\uparrow$} 
        & {\cellcolor{gray!40}$S_\alpha$$\uparrow$}
        & {\cellcolor{gray!40}$M$$\downarrow$} 
        & {\cellcolor{gray!40}$F^\omega_\beta$$\uparrow$} 
        & {\cellcolor{gray!40}$E_\phi$$\uparrow$} 
        & {\cellcolor{gray!40}$S_\alpha$$\uparrow$}
        & {\cellcolor{gray!40}$M$$\downarrow$} 
        & {\cellcolor{gray!40}$F^\omega_\beta$$\uparrow$} 
        & {\cellcolor{gray!40}$E_\phi$$\uparrow$} 
        & {\cellcolor{gray!40}$S_\alpha$$\uparrow$}
        & {\cellcolor{gray!40}$M$$\downarrow$} 
        & {\cellcolor{gray!40}$F^\omega_\beta$$\uparrow$} 
        & {\cellcolor{gray!40}$E_\phi$$\uparrow$} 
        & {\cellcolor{gray!40}$S_\alpha$$\uparrow$} \\
        \midrule

        \multicolumn{17}{c}{CNN-based Methods (ResNet50)} \\
        \midrule
        SINet~\cite{fan2020CamouflageSINet} 
        & 0.034 & 0.798 & 0.936 & 0.873 
        & 0.092 & 0.670 & 0.825 & 0.746 
        & 0.043 & 0.631 & 0.865 & 0.778 
        & 0.058 & 0.737 & 0.882 & 0.809 \\
        
        \rowcolor{gray!30}
        \textit{+UMBD} 
        & 0.029 & 0.825 & 0.960 & 0.883 
        & 0.084 & 0.711 & 0.831 & 0.764 
        & 0.040 & 0.674 & 0.876 & 0.786 
        & 0.054 & 0.772 & 0.885 & 0.818 \\
        
        FEDER~\cite{he2023camouflaged_feder} 
        & 0.030 & 0.824 & 0.941 & 0.888 
        & 0.073 & 0.740 & 0.866 & 0.799 
        & 0.032 & 0.713 & 0.899 & 0.822 
        & 0.045 & 0.796 & 0.909 & 0.847 \\
        
        \rowcolor{gray!30}
        \textit{+UMBD} 
        & 0.028 & 0.838 & 0.949 & 0.890 
        & 0.069 & 0.757 & 0.871 & 0.807 
        & 0.030 & 0.732 & 0.905 & 0.827 
        & 0.043 & 0.809 & 0.912 & 0.849 \\

\textbf{Average Improve(\%)} & \textbf{10.96} & \textbf{2.54} & \textbf{1.71} & \textbf{0.69} & \textbf{7.09} & \textbf{4.21} & \textbf{0.65} & \textbf{1.71} & \textbf{6.61} & \textbf{4.74} & \textbf{0.97} & \textbf{0.82} & \textbf{5.67} & \textbf{3.19} & \textbf{0.34} & \textbf{0.67} \\
        \midrule

        \multicolumn{17}{c}{CNN-based Methods (Res2Net50)} \\
        \midrule
        SINetV2~\cite{fan2021concealed_survey0} 
        & 0.029 & 0.792 & 0.922 & 0.890 
        & 0.071 & 0.733 & 0.875 & 0.822 
        & 0.036 & 0.668 & 0.867 & 0.820 
        & 0.048 & 0.769 & 0.898 & 0.848 \\
        
        \rowcolor{gray!30}
        \textit{+UMBD} 
        & 0.027 & 0.853 & 0.955 & 0.892 
        & 0.067 & 0.774 & 0.885 & 0.829 
        & 0.032 & 0.733 & 0.909 & 0.825 
        & 0.044 & 0.807 & 0.911 & 0.849 \\
        
        MRRNet~\cite{yan2023camouflagedMRRNet} 
        & 0.031 & 0.794 & 0.923 & 0.889 
        & 0.069 & 0.750 & 0.882 & 0.829 
        & 0.034 & 0.692 & 0.878 & 0.831 
        & 0.046 & 0.776 & 0.898 & 0.855 \\
        
        \rowcolor{gray!30}
        \textit{+UMBD} 
        & 0.030 & 0.796 & 0.946 & 0.894 
        & 0.068 & 0.756 & 0.893 & 0.830 
        & 0.033 & 0.695 & 0.904 & 0.835 
        & 0.045 & 0.784 & 0.908 & 0.862 \\
        
\textbf{Average Improve(\%)} & \textbf{5.06} & \textbf{3.98} & \textbf{3.04} & \textbf{0.39} & \textbf{3.54} & \textbf{3.20} & \textbf{1.20} & \textbf{0.49} & \textbf{7.03} & \textbf{5.08} & \textbf{3.90} & \textbf{0.55} & \textbf{5.25} & \textbf{2.99} & \textbf{1.28} & \textbf{0.47} \\
        \midrule

        \multicolumn{17}{c}{Transformer-based Methods (PVTv2)} \\
        \midrule
        Camoformer~\cite{yin2024camoformer} 
        & 0.026 & 0.842 & 0.948 & 0.898 
        & 0.046 & 0.819 & 0.920 & 0.873 
        & 0.025 & 0.763 & 0.917 & 0.861 
        & 0.032 & 0.837 & 0.927 & 0.887 \\
        
        \rowcolor{gray!30}
        \textit{+UMBD} 
        & 0.025 & 0.853 & 0.956 & 0.898 
        & 0.044 & 0.824 & 0.922 & 0.875  
        & 0.024 & 0.773 & 0.926 & 0.862 
        & 0.031 & 0.843 & 0.929 & 0.888 \\
        
        FSNet~\cite{song2023fsnet} 
        & 0.023 & 0.837 & 0.950 & 0.905 
        & 0.042 & 0.827 & 0.928 & 0.880 
        & 0.023 & 0.776 & 0.931 & 0.870 
        & 0.031 & 0.843 & 0.933 & 0.892 \\
        
        \rowcolor{gray!30}
        \textit{+UMBD} 
        & 0.022 & 0.849 & 0.960 & 0.909 
        & 0.041 & 0.838 & 0.929 & 0.880  
        & 0.022 & 0.783 & 0.936 & 0.872 
        & 0.030 & 0.852 & 0.935 & 0.893 \\
        
\textbf{Average Improve(\%)} & \textbf{4.10} & \textbf{1.37} & \textbf{0.95} & \textbf{0.22} & \textbf{3.36} & \textbf{0.97} & \textbf{0.16} & \textbf{0.11} & \textbf{4.17} & \textbf{1.11} & \textbf{0.76} & \textbf{0.17} & \textbf{3.18} & \textbf{0.89} & \textbf{0.22} & \textbf{0.11} \\
        \bottomrule
    \end{tabular}}
      \vspace{-3mm}
\end{table*}

\begin{figure}[!t]
  \centering
  \includegraphics[width=1.0 \textwidth]{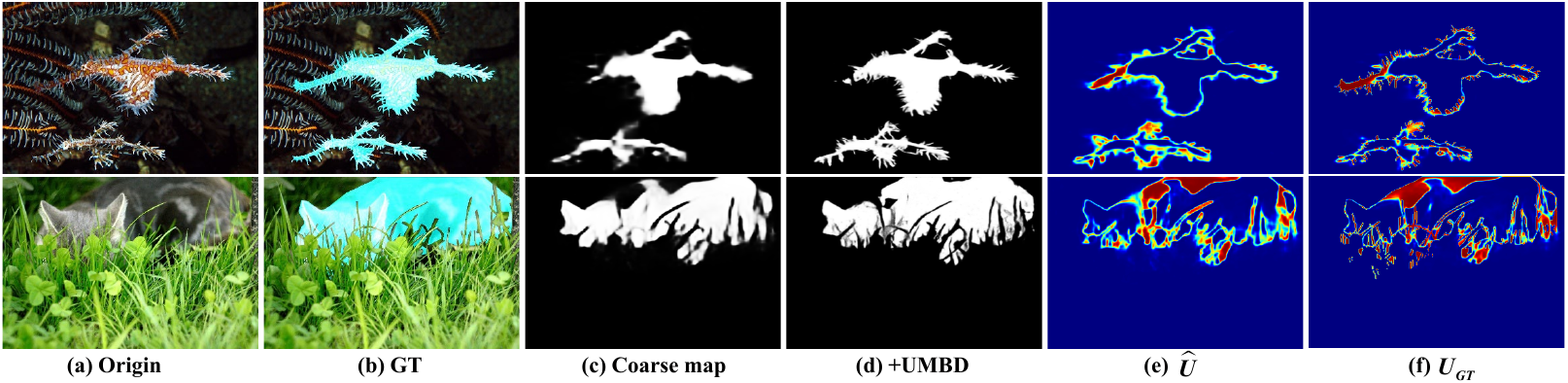}
  \vspace{-5mm}
  \caption{Qualitative refinement results by our UMBD with uncertainty estimates $\hat{U}$ versus ground truth $U_{GT}$. Samples are derived from SINet and FEDER. Zoom in for a better view.
  }
\label{fig:methods_refinement_with_U}
  \vspace{-3mm}
\end{figure}
\parhead{Refinement Gains.}
As shown in Tab.~\ref{table:Results_COD}, our UMBD framework achieves significant refinement gains and demonstrates universal compatibility with various existing COD methods across three backbones: ResNet50~\cite{he2016deepresnet50}, Res2Net50~\cite{gao2019res2net}, and PVTv2~\cite{wang2022pvt}. Notably, it attains average improvements of 5.5\% in MAE and 3.2\% in $F^\omega_\beta$. Even for Transformer-based SOTA methods, our approach still delivers measurable performance enhancements, underscoring the effectiveness and generalization capabilities of our UMBD framework. 
As illustrated in Figs.~\ref{fig:Visual_comparison_with_barchart} and \ref{fig:methods_refinement_with_U}, our method significantly mitigates errors, including blurred edges, missing details, and significant false positives/negatives. Columns (e)-(f) in Fig.~\ref{fig:methods_refinement_with_U} further demonstrate HUQNet's uncertainty map estimations $\hat{U}$ and the corresponding ground truth. HUQNet accurately estimates $U_{GT}$, not only in areas with ambiguous initial segmentation results but also by identifying large regions overlooked by existing methods, thereby providing robust uncertainty masking for the refinement process.

\begin{table*}[!b]
    \setlength{\abovecaptionskip}{0cm}
    \setlength{\belowcaptionskip}{-0.2cm}
    \centering
    \caption{Quantitative comparison of recent segmentation refiners and our proposed UMBD. The best two results from refiners are in {\color[HTML]{FF0000} \textbf{red}} and {\color[HTML]{00B0F0} \textbf{blue}} fonts.}
    \label{tab:refiner_comparison}
    \resizebox{\linewidth}{!}{
    \setlength{\tabcolsep}{1.4mm}
    \begin{tabular}{l|cccc|cccc|cccc|cccc}
        \toprule
        \multicolumn{1}{c|}{\multirow{2}{*}{Methods}} 
        & \multicolumn{4}{c|}{\textit{CHAMELEON}} 
        & \multicolumn{4}{c|}{\textit{CAMO}} 
        & \multicolumn{4}{c|}{\textit{COD10K}} 
        & \multicolumn{4}{c}{\textit{NC4K}} \\ 
        \cline{2-17}
        & {\cellcolor{gray!40}$M$$\downarrow$} 
        & {\cellcolor{gray!40}$F^\omega_\beta$$\uparrow$} 
        & {\cellcolor{gray!40}$E_\phi$$\uparrow$} 
        & {\cellcolor{gray!40}$S_\alpha$$\uparrow$}
        & {\cellcolor{gray!40}$M$$\downarrow$} 
        & {\cellcolor{gray!40}$F^\omega_\beta$$\uparrow$} 
        & {\cellcolor{gray!40}$E_\phi$$\uparrow$} 
        & {\cellcolor{gray!40}$S_\alpha$$\uparrow$}
        & {\cellcolor{gray!40}$M$$\downarrow$} 
        & {\cellcolor{gray!40}$F^\omega_\beta$$\uparrow$} 
        & {\cellcolor{gray!40}$E_\phi$$\uparrow$} 
        & {\cellcolor{gray!40}$S_\alpha$$\uparrow$}
        & {\cellcolor{gray!40}$M$$\downarrow$} 
        & {\cellcolor{gray!40}$F^\omega_\beta$$\uparrow$} 
        & {\cellcolor{gray!40}$E_\phi$$\uparrow$} 
        & {\cellcolor{gray!40}$S_\alpha$$\uparrow$} \\
        \midrule
        
        SINetV2~\cite{fan2021concealed_survey0} 
        & 0.029 & 0.792 & 0.922 & 0.890 
        & 0.071 & 0.733 & 0.875 & 0.822 
        & 0.036 & 0.668 & 0.867 & 0.820 
        & 0.048 & 0.769 & 0.898 & 0.848 \\
        
        \textit{+Hidiff}~\cite{chen2024hidiff}
        & 0.062 & 0.581 & 0.838 & 0.844 
        & 0.099 & 0.571 & 0.820 & {\color[HTML]{00B0F0}\textbf{0.786}} 
        & 0.062 & 0.434 & 0.700 & 0.758 
        & 0.084 & 0.540 & 0.797 & 0.792 \\ 
        
        \textit{+SegRefiner}~\cite{wang2023segrefiner}
        & 0.073 & 0.619 & 0.812 & 0.777 
        & 0.116 & 0.583 & 0.763 & 0.736 
        & 0.079 & 0.487 & 0.764 & 0.714 
        & 0.075 & 0.635 & 0.830 & 0.781 \\ 
        
        \textit{+SAMRefiner}~\cite{lin2025samrefiner}
        & {\color[HTML]{00B0F0}\textbf{0.029}} & {\color[HTML]{FF0000}\textbf{0.872}} & {\color[HTML]{00B0F0}\textbf{0.906}} & {\color[HTML]{00B0F0}\textbf{0.857}} 
        & {\color[HTML]{00B0F0}\textbf{0.077}} & {\color[HTML]{FF0000}\textbf{0.785}} & {\color[HTML]{00B0F0}\textbf{0.828}} & 0.776
        & {\color[HTML]{00B0F0}\textbf{0.033}} &{\color[HTML]{FF0000}\textbf{ 0.784}} &{\color[HTML]{00B0F0}\textbf{ 0.867}} & {\color[HTML]{00B0F0}\textbf{0.813}}
        & {\color[HTML]{00B0F0}\textbf{0.048}} &{\color[HTML]{FF0000}\textbf{ 0.836}} & {\color[HTML]{00B0F0}\textbf{0.877}} & {\color[HTML]{00B0F0}\textbf{0.831}}
         \\

        \textit{+UMBD}
        & {\color[HTML]{FF0000}\textbf{0.027}} & {\color[HTML]{00B0F0}\textbf{0.853}} & {\color[HTML]{FF0000}\textbf{0.955}} &{\color[HTML]{FF0000}\textbf{ 0.892}} 
        & {\color[HTML]{FF0000}\textbf{0.067}} &{\color[HTML]{00B0F0}\textbf{ 0.774}} & {\color[HTML]{FF0000}\textbf{0.885}} & {\color[HTML]{FF0000}\textbf{0.829 }}
        & {\color[HTML]{FF0000}\textbf{0.032}} & {\color[HTML]{00B0F0}\textbf{0.733}} & {\color[HTML]{FF0000}\textbf{0.909}} & {\color[HTML]{FF0000}\textbf{0.825 }}
        & {\color[HTML]{FF0000}\textbf{0.044 }}& {\color[HTML]{00B0F0}\textbf{0.807}} & {\color[HTML]{FF0000}\textbf{0.911}} &{\color[HTML]{FF0000}\textbf{ 0.849}}
         \\
        \midrule
        
        FEDER~\cite{he2023camouflaged_feder} 
        & 0.030 & 0.824 & 0.941 & 0.888 
        & 0.073 & 0.740 & 0.866 & 0.799 
        & 0.032 & 0.713 & 0.899 & 0.822 
        & 0.045 & 0.796 & 0.909 & 0.847 \\
        
        \textit{+Hidiff}~\cite{chen2024hidiff}
        & 0.070 & 0.554 & 0.814 & 0.825 
        & 0.111 & 0.528 & 0.789 & 0.753 
        & 0.070 & 0.389 & 0.665 & 0.727 
        & 0.088 & 0.513 & 0.774 & 0.773 \\
        
        \textit{+SegRefiner}~\cite{wang2023segrefiner} 
        & 0.041 & 0.751 & 0.908 & 0.850 
        & 0.081 & 0.683 & {\color[HTML]{00B0F0}\textbf{0.848}} & {\color[HTML]{00B0F0}\textbf{0.785 }}
        & 0.043 & 0.635 & 0.866 & 0.794 
        & 0.050 & 0.742 & {\color[HTML]{00B0F0}\textbf{0.896}} & 0.832  \\
        
        \textit{+SAMRefiner}~\cite{lin2025samrefiner}
        & {\color[HTML]{00B0F0}\textbf{0.035}} & {\color[HTML]{FF0000}\textbf{0.859}} & {\color[HTML]{00B0F0}\textbf{0.919}} & {\color[HTML]{00B0F0}\textbf{0.854}}
        & {\color[HTML]{00B0F0}\textbf{0.082}} & {\color[HTML]{FF0000}\textbf{0.769}} & 0.809 & 0.768
        & {\color[HTML]{00B0F0}\textbf{0.031}} & {\color[HTML]{FF0000}\textbf{0.792}} & {\color[HTML]{00B0F0}\textbf{0.881}} & {\color[HTML]{00B0F0}\textbf{0.823}} 
        & {\color[HTML]{00B0F0}\textbf{0.047}} & {\color[HTML]{FF0000}\textbf{0.838}} & 0.880 & {\color[HTML]{00B0F0}\textbf{0.833}} \\

        \textit{+UMBD}
        & {\color[HTML]{FF0000} \textbf{0.028}} &{\color[HTML]{00B0F0} \textbf{ 0.838}} & {\color[HTML]{FF0000} \textbf{0.949}} & {\color[HTML]{FF0000} \textbf{0.890}} 
        & {\color[HTML]{FF0000} \textbf{0.069}} & {\color[HTML]{00B0F0} \textbf{0.757}} & {\color[HTML]{FF0000} \textbf{0.871}} & {\color[HTML]{FF0000} \textbf{0.807} }
        & {\color[HTML]{FF0000} \textbf{0.030}} & {\color[HTML]{00B0F0} \textbf{0.732}} & {\color[HTML]{FF0000} \textbf{0.905}} & {\color[HTML]{FF0000} \textbf{0.827 }}
        & {\color[HTML]{FF0000} \textbf{0.043 }}& {\color[HTML]{00B0F0} \textbf{0.809}} & {\color[HTML]{FF0000} \textbf{0.912}} & {\color[HTML]{FF0000} \textbf{0.849}}\\
        \bottomrule
    \end{tabular}}
\end{table*}

\begin{figure}[!b]
  \centering
  \includegraphics[width=1.0\linewidth]{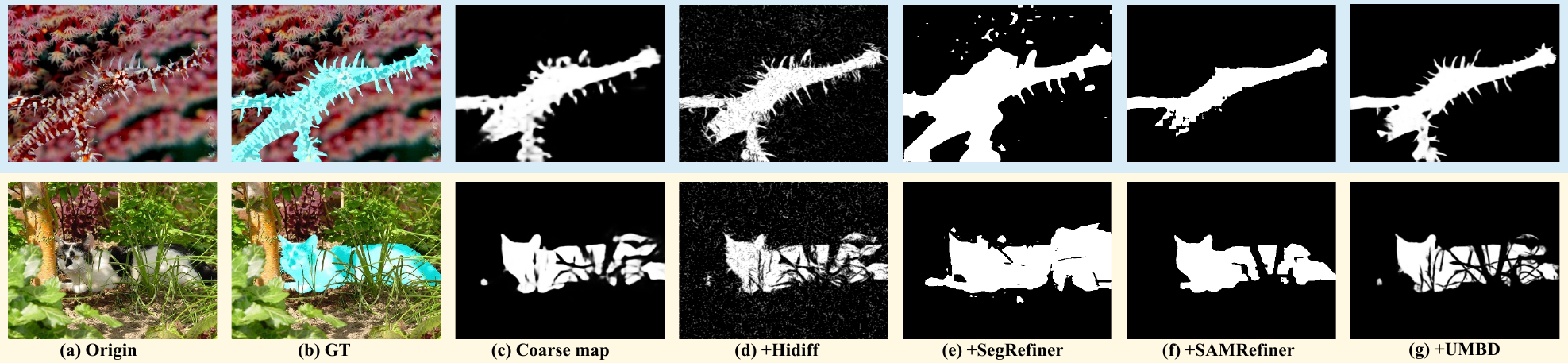}
  \vspace{-3mm}
  \caption{Qualitative comparisons with other refiners on COD benchmarks. Blue/yellow backgrounds indicate refined outputs from SINetV2 and FEDER respectively. Please zoom in for a better view.}
\label{fig:Visual_comparison_with_different_refiner}
  \vspace{-3mm}
\end{figure}

\parhead{Comparison with Existing Refiners.}\label{sec:Comparison with Existing Refiners}
We evaluate the proposed UMBD against three highly versatile SOTA refinement methods: Hidiff~\cite{chen2024hidiff}, Segrefiner~\cite{wang2023segrefiner} (the only two existing diffusion-based refiners) and SAMRefiner~\cite{lin2025samrefiner} (model-agnostic SOTA method). Comprehensive evaluations using SINetV2~\cite{fan2021concealed_survey0} and FEDER~\cite{he2023camouflaged_feder} as prior segmenters demonstrate our method's superior performance across most metrics, as shown in Tab.~\ref{tab:refiner_comparison}.
Notably, while SAMRefiner achieves higher $F^\omega_\beta$ scores, it underperforms the coarse masks on three other metrics and exhibits structural/detail degradation (see in Fig.~\ref{fig:Visual_comparison_with_different_refiner}), for its SAM foundation is not well-suited for COD~\cite{chen2023sam-adapter}. In fact, the intrinsic complexity of COD tasks renders refinement particularly challenging, and our UMBD is emerging as the only method that achieves stable improvements.
Critically, existing diffusion-based refiners fail to focus on modeling residual regions via uncertainty-aware mechanisms. Given the high foreground-background similarity in COD tasks, global noise learning inadvertently introduces uncertainty into correctly segmented areas, leading to either incomplete denoising or error propagation in missegmented regions (see Fig.~\ref{fig:Visual_comparison_with_different_refiner} (d, e)). More qualitative results can be found in Fig.~\ref{fig:Visual_comparison_with_different_refiner_left} (Appx.~\ref{more_res}).

\subsection{Ablation Study}\label{sec:AblationStudy}

We conduct ablation studies on COD10K with pretrained SINetV2 as  the prior segmenter.

\parhead{Effect of HUQNet.}
As shown in Tab.~\ref{table:HUQNet}, disabling the uncertainty mask mechanism in our UMBD refiner leads to incomplete noise removal similar to Hidiff~\cite{chen2024hidiff}, validating its critical role. Replacing the proposed HUQNet’s output with a roughly obtained uncertainty mask (\eg, $U_E$) from non-learnable methods still achieves effective refinement over $M_c$, demonstrating the robustness of our framework. Performance degradation observed when removing any HUQNet module underscores the necessity of these components. 
Notably, without the RAM module that injects prior knowledge across feature stages, the basic Encoder-Decoder architecture fails to provide effective estimation on challenging COD, failing to to yield refined results.
Furthermore, ablating the entire BNN branch versus retaining it without utilizing $U_B$ reveals that even without explicit integration of $U_B$, the shared encoder enables mutual enhancement between BNN and discriminative branches.

\begin{table*}[htb]
\centering
\setlength{\abovecaptionskip}{0.1cm}
\begin{minipage}{.67\linewidth}
    \centering
    \setlength{\abovecaptionskip}{0cm}
    \caption{Effect of uncertainty mask mechanism and HUQNet.}
    \resizebox{\columnwidth}{!}{
        \setlength{\tabcolsep}{1mm}
        \begin{tabular}{c|c|c|c|ccccc|c}
    \toprule
    \multirow{2}{*}{Metrics} 
    & \multirow{2}{*}{$M_c$}  & w/o  
    & $\hat{U}=$ 
    & \multicolumn{5}{c|}{Effect of HUQNet} 
    &\cellcolor{c2!20}{UMBD} \\ 
    \cline{5-9}
    &  & $\hat{U}$
    & $U_E$
    & w/o BNN  
    & w/o $U_B$ 
    & w/o $U_E$ 
    & w/o CA
    & \multicolumn{1}{c|}{w/o RAM}    
    &\cellcolor{c2!20}{(Ours)} \\ 
    \midrule
    $M~\downarrow$  & 0.036 & 0.062 & 0.035 & 0.034 & 0.033 & 0.035 &0.034 &0.039 &\cellcolor{c2!20}\textbf{0.032} \\
    $F^\omega_\beta~\uparrow$ & 0.668 & 0.434 & 0.713 & 0.715 & 0.725 & 0.721 &0.720 &0.659 &\cellcolor{c2!20}\textbf{0.733} \\
    $E_\phi~\uparrow$ & 0.867 & 0.700 & 0.900 & 0.902 & 0.906 & 0.905 &0.903 &0.865 &\cellcolor{c2!20}\textbf{0.909} \\
    $S_\alpha~\uparrow$ & 0.820 & 0.758  & 0.821 & 0.823 & 0.824 & 0.824 &0.822 &0.815 &\cellcolor{c2!20}\textbf{0.825} \\
    \bottomrule
\end{tabular}\label{table:HUQNet}
    }
\end{minipage}
\begin{minipage}{.32\linewidth}
    \centering
    \setlength{\abovecaptionskip}{0cm}
    \caption{Effect of kernel choice.}
    \resizebox{\columnwidth}{!}{
        \setlength{\tabcolsep}{1mm}
        \begin{tabular}{c|c|c|c}
            \toprule
            \multirow{2}{*}{Metrics} 
             & \multirow{2}{*}{$M_c$}
            & Gaussian
            & Bernoulli
             \\ &
            & kernel
            & kernel \\ 
            \midrule
            $M~\downarrow$  & 0.036 & 0.034 & 0.032 \\
            $F^\omega_\beta~\uparrow$ & 0.668 & 0.701 & 0.733 \\
            $E_\phi~\uparrow$ & 0.867 & 0.881 & 0.909  \\
            $S_\alpha~\uparrow$ & 0.820 & 0.822  & 0.825  \\
            \bottomrule
    \end{tabular}\label{table:kernel_choice}
    }
\end{minipage}
\end{table*}

\parhead{Effect of Bernoulli Diffusion.} 
As shown in Tab.~\ref{table:kernel_choice}, substituting the Bernoulli kernel in our diffusion model with a Gaussian kernel leads to performance degradation, validating the superior suitability of Bernoulli diffusion models for segmentation refinement. Benefiting from our residual modeling strategy, the diffusion process focuses exclusively on poorly segmented regions instead of modeling the global segmentation mask from scratch. 
Consequently, as shown in Tab.~\ref{table:abl_time}, our UMBD yields robust refinement even with a reduced number of inference steps (\eg, $T=3$), comprehensive efficiency analysis can be found in Appx.~\ref{sec:Efficiency Analysis}. While we follow the standard setting of $T=10$~\cite{chen2024camodiffusion,yang2025uncertaintyUGDM} in Sec.~\ref{sec:Implementation_Details}, our framework allows dynamic adjustment of sampling steps based on the required refinement quality in practice, which effectively mitigates the heavy computational overhead commonly associated with diffusion-based methods. The refinement time is measured at a resolution of $384 \times 384$ using an RTX 4090 GPU.

\begin{table*}[htb]
\centering
\setlength{\abovecaptionskip}{0cm}
\caption{Ablation study on sampling steps $T$. Refinement time measured on an RTX 4090}
\resizebox{1.0\linewidth}{!}{
\setlength{\tabcolsep}{1mm}
\begin{tabular}{c|c|c|c|c|c|c|c|c|c|c|c}
\toprule
Metrics &SINetV2 & $T=1$ & $T=2$ &$ T=3$ & $T=4 $& $T=5$ & $T=6$ & $T=7$ & $T=8$ & $T=9$ & $T=10$ \\ 
\midrule
$M~\downarrow$  & 0.036 & 0.067 & 0.035 & 0.033 & 0.033 & 0.033 & 0.033 & 0.032 & 0.032 & 0.032 & 0.032 \\
$F^\omega_\beta~\uparrow$ & 0.668& 0.491 & 0.697 & 0.714 & 0.721 & 0.725 & 0.725 & 0.732 & 0.731 & 0.735 & 0.733 \\
$E_\phi~\uparrow$& 0.867 & 0.856 & 0.885 & 0.896 & 0.902 & 0.906 & 0.907 & 0.909 & 0.909 & 0.910 & 0.909 \\
$S_\alpha~\uparrow$ & 0.820 & 0.658 & 0.817 & 0.826 & 0.826 & 0.824 & 0.824 & 0.823 & 0.822 & 0.820 & 0.825 \\
    
t[s/img] & \cellcolor{gray!40}- & \cellcolor{gray!40}0.072 & \cellcolor{gray!40}0.092 & \cellcolor{gray!40}0.112 & \cellcolor{gray!40}0.133 & \cellcolor{gray!40}0.160 & \cellcolor{gray!40}0.177 & \cellcolor{gray!40}0.192 & \cellcolor{gray!40}0.213 & \cellcolor{gray!40}0.228 & \cellcolor{gray!40}0.251 \\
\bottomrule
\end{tabular}
}
\label{table:abl_time}
\end{table*}

\section{Conclusions}\label{sec:Conclusion}
We present the UMBD framework, the first generative refinement solution designed for COD. By strategically integrating Bernoulli diffusion models with a novel uncertainty-masked mechanism through our HUQNet, our approach enables precise residual refinement while preserving accurate segmentation regions. The framework demonstrates strong compatibility with various existing methods, achieving remarkable refinement performance with modest computational overhead. The inherent plug-and-play structure underscores the effectiveness and adaptability of the proposed refinement framework not only for COD, but also potentially for other vision segmentation tasks.

\parhead{Limitations and Future Work.}\label{sec:limitations}
While our current implementation employs DDIM acceleration to balance refinement quality and computational efficiency, alternative sampling strategies~\cite{lu2022dpm} worth exploring. Moreover, temporal aggregation mechanisms~\cite{chen2024camodiffusion} that fuse multi-step diffusion predictions may further improve detail recovery.

\small
\bibliographystyle{IEEEtran}
\bibliography{refs.bib}


\appendix
\section{Appendix}


\subsection{Efficiency Analysis}\label{sec:Efficiency Analysis}

\begin{table*}[!h]
    \setlength{\abovecaptionskip}{0cm}
    \setlength{\belowcaptionskip}{-0.2cm}
    \centering
    \caption{Impact of reducing steps $T$ on refinement performance. The best results are marked in \textbf{bold}.}
    \label{table:Results_Efficiency_Analysis}
    \vspace{1mm}
    \resizebox{\textwidth}{!}{%
        \setlength{\tabcolsep}{1.4mm}
        \begin{tabular}{l|cccc|cccc|cccc|cccc}
            \toprule
            \multicolumn{1}{c|}{\multirow{2}{*}{Methods}}& \multicolumn{4}{c|}{\textit{CHAMELEON}} & \multicolumn{4}{c|}{\textit{CAMO}} & \multicolumn{4}{c|}{\textit{COD10K}} & \multicolumn{4}{c}{\textit{NC4K}} \\ 
            \cline{2-17}
            & {\cellcolor{gray!40}$M$$\downarrow$} 
            & {\cellcolor{gray!40}$F^\omega_\beta$$\uparrow$} 
            & {\cellcolor{gray!40}$E_\phi$$\uparrow$} 
            & {\cellcolor{gray!40}$S_\alpha$$\uparrow$}
            & {\cellcolor{gray!40}$M$$\downarrow$} 
            & {\cellcolor{gray!40}$F^\omega_\beta$$\uparrow$} 
            & {\cellcolor{gray!40}$E_\phi$$\uparrow$} 
            & {\cellcolor{gray!40}$S_\alpha$$\uparrow$}
            & {\cellcolor{gray!40}$M$$\downarrow$} 
            & {\cellcolor{gray!40}$F^\omega_\beta$$\uparrow$} 
            & {\cellcolor{gray!40}$E_\phi$$\uparrow$} 
            & {\cellcolor{gray!40}$S_\alpha$$\uparrow$}
            & {\cellcolor{gray!40}$M$$\downarrow$} 
            & {\cellcolor{gray!40}$F^\omega_\beta$$\uparrow$} 
            & {\cellcolor{gray!40}$E_\phi$$\uparrow$} 
            & {\cellcolor{gray!40}$S_\alpha$$\uparrow$} \\
            \midrule
            SINetV2~\cite{fan2021concealed_survey0} 
            & 0.029 & 0.792 & 0.922 & 0.890 
            & 0.071 & 0.733 & 0.875 & 0.822 
            & 0.036 & 0.668 & 0.867 & 0.820 
            & 0.048 & 0.769 & 0.898 & 0.848 \\
            +UMBD(T=2) & 0.030 & 0.814 & 0.938 & 0.880 & 0.073 & 0.741 & 0.876 & 0.803 & 0.035 & 0.697 & 0.885 & 0.817 & 0.049 & 0.783 & 0.902 & 0.837 \\
            +UMBD(T=3) & \textbf{0.027} & 0.831 & 0.947 & \textbf{0.894} & 0.069 & 0.759 & 0.883 & 0.815 & 0.033 & 0.714 & 0.896 & 0.826 & 0.046 & 0.796 & 0.907 & 0.846 \\
            +UMBD(T=5) & \textbf{0.027} & 0.840 & \textbf{0.955} & \textbf{0.894} & 0.068 & 0.768 & \textbf{0.887} & 0.815 & 0.033 & 0.725 & 0.906 & 0.824 & 0.045 & 0.803 & 0.910 & 0.846 \\
            +UMBD(T=10)
            & \textbf{0.027} & \textbf{0.853} & \textbf{0.955} & 0.892 
            & \textbf{0.067} & \textbf{0.774} & 0.885 & \textbf{0.829} 
            & \textbf{0.032} & \textbf{0.733} & \textbf{0.909} & \textbf{0.825} 
            & \textbf{0.044} & \textbf{0.807} & \textbf{0.911} & \textbf{0.849} \\
            \midrule 
            FEDER~\cite{he2023camouflaged_feder} 
            & 0.030 & 0.824 & 0.941 & 0.888 
            & 0.073 & 0.740 & 0.866 & 0.799 
            & 0.032 & 0.713 & 0.899 & 0.822 
            & 0.045 & 0.796 & 0.909 & 0.847 \\
            +UMBD(T=2) & 0.030 & 0.836 & 0.947 & 0.890 & 0.072 & 0.748 & 0.866 & 0.802 & 0.031 & 0.722 & 0.900 & 0.825 & 0.044 & 0.804 & 0.910 & 0.847 \\
            +UMBD(T=3) & 0.029 & 0.837 & 0.948 & 0.891 & 0.071 & 0.752 & 0.868 & 0.804 & 0.031 & 0.727 & 0.902 & \textbf{0.827} & 0.044 & 0.807 & \textbf{0.912} & \textbf{0.849} \\
            +UMBD(T=5) & 0.029 & 0.837 & \textbf{0.949} & \textbf{0.891} & 0.070 & 0.755 & 0.869 & 0.805 & 0.031 & 0.730 & 0.904 & \textbf{0.827} & \textbf{0.043} & \textbf{0.809} & \textbf{0.912} & \textbf{0.849} \\
            +UMBD(T=10)
            & \textbf{0.028} & \textbf{0.838} & \textbf{0.949} & 0.890 
            & \textbf{0.069} & \textbf{0.757} & \textbf{0.871} & \textbf{0.807} 
            & \textbf{0.030} & \textbf{0.732} & \textbf{0.905} & \textbf{0.827} 
            & \textbf{0.043} & \textbf{0.809} & \textbf{0.912} & \textbf{0.849} \\
            \bottomrule
        \end{tabular}%
    }
\end{table*}

As comprehensively demonstrated in Tab.~\ref{table:Results_Efficiency_Analysis}, the refinement gains exhibit negligible degradation when reducing the sampling steps 
$T$ from 10 to 3. This phenomenon occurs because our residual modeling mechanism effectively preserves the originally correct segmentation regions. While decreasing 
$T$ within this range may introduce minor quality impacts on refined areas, it prevents catastrophic collapse of the overall segmentation quality, which is a significant efficiency improvement compared to other diffusion-based segmentation methods~\cite{chen2024hidiff,wang2023segrefiner,chen2024camodiffusion,yang2025uncertaintyUGDM}.

\subsection{Extended Experiments Beyond COD}\label{Extended Experiments}
To demonstrate the generalization capability of the proposed UMBD refinement framework across different segmentation tasks, we also evaluate its performance on polyp segmentation~\cite{mei2025surveypoly} and transparent object detection~\cite{tong2024confusing}, which together with COD, falls under the umbrella of Concealed Object Segmentation (COS). This extension validates the framework’s applicability to broader scenarios.

\subsubsection{Experiments on Polyp Segmentation}\label{res_poly}

\parhead{Datasets and Metrics.} For polyp segmentation, we utilize two benchmarks: \textit{CVC-300}~\cite{bernal2015wmCVC300} and \textit{ETIS}~\cite{silva2014towardETIS}. The training protocol follows the setup of PraNet~\cite{fan2020pranet}. Quantitative evaluation is conducted using three commonly adopted metrics: mean absolute error ($M$), mean Intersection over Union (mIoU), and adaptive F-measure ($F_\beta$). Superior performance is indicated by lower values of $M$ and higher values of $F_\beta$ and $\text{mIoU}$.

\begin{figure}[!h]
  \centering
  \includegraphics[width=1.0 \textwidth]{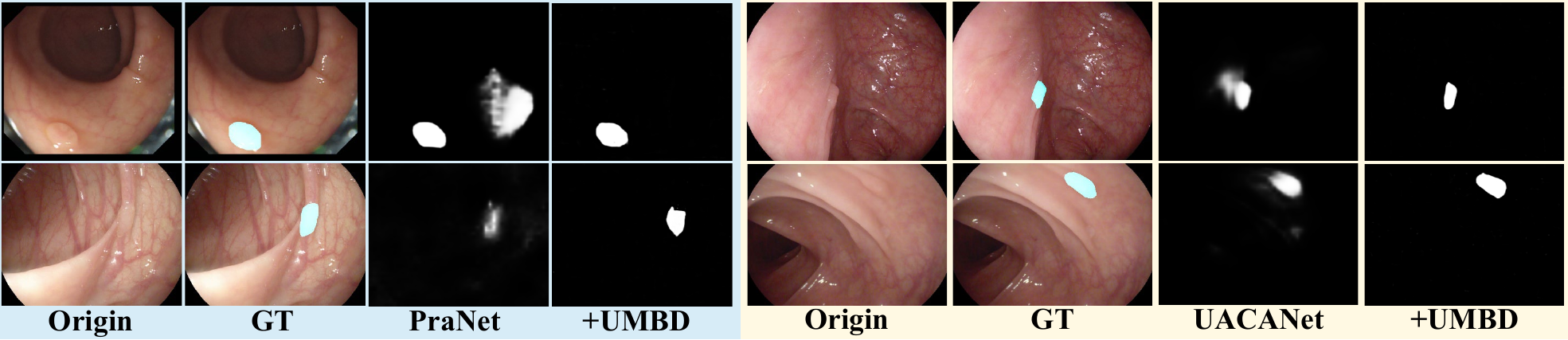}
  \vspace{-5mm}
  \caption{Qualitative results by our UMBD on polyp segmentation. Please zoom in for a better view.
  }
\label{fig:methods_refinement_poly}
  \vspace{-3mm}
\end{figure}

\parhead{Results and Comparative Evaluation.} As demonstrated by the quantitative results in Tab.~\ref{table:Results_CVC_ETIS} and qualitative visualization in Fig.~\ref{fig:methods_refinement_poly}, our UMBD framework achieves substantial refinement gains in polyp segmentation results from PraNet~\cite{fan2020pranet} and UACANet~\cite{Kim2021UACANetUA}, which are even more pronounced than those observed in COD tasks. The visual evidence confirms the method's capability in effectively addressing blurred predictions, large-area false positives/negatives, and structural refinement of small targets.

\begin{table*}[!h]
    \setlength{\abovecaptionskip}{0cm}
    \setlength{\belowcaptionskip}{-0.2cm}
    \centering
    \caption{Quantitative performance gains on polyp segmentation. Average improvement rates (\%) are marked in \textbf{bold}.}
    \label{table:Results_CVC_ETIS}
    \vspace{1mm}
    \resizebox{0.7\textwidth}{!}{
    \setlength{\tabcolsep}{2.5mm}
    \begin{tabular}{l|ccc|ccc}
        \toprule
        \multicolumn{1}{c|}{\multirow{2}{*}{Methods}} 
        & \multicolumn{3}{c|}{\textit{CVC-300}} 
        & \multicolumn{3}{c}{\textit{ETIS}} \\
        \cline{2-7}
        & {\cellcolor{gray!40}$M$$\downarrow$} 
        & {\cellcolor{gray!40}$F_\beta$$\uparrow$} 
        & {\cellcolor{gray!40}$\text{mIoU}$$\uparrow$}
        & {\cellcolor{gray!40}$M$$\downarrow$} 
        & {\cellcolor{gray!40}$F_\beta$$\uparrow$} 
        & {\cellcolor{gray!40}$\text{mIoU}$$\uparrow$} \\
        \midrule

        PraNet~\cite{fan2020pranet} & 0.010 & 0.820 & 0.804 & 0.031 & 0.600 & 0.576 \\
        \rowcolor{gray!30}
        \textit{+UMBD} & 0.006 & 0.880 & 0.840 & 0.015 & 0.616 & 0.586 \\
        UACANet-S~\cite{Kim2021UACANetUA} & 0.007 & 0.809 & 0.841 & 0.026 & 0.550 & 0.619 \\
        \rowcolor{gray!30}
        \textit{+UMBD} & 0.006 & 0.856 & 0.845 & 0.016 & 0.615 & 0.625 \\
        \midrule
        
        \textbf{Average Improve (\%)}& \textbf{27.15} & \textbf{6.56} & \textbf{2.48} & \textbf{45.04} & \textbf{7.24} & \textbf{1.35} \\
        \bottomrule
    \end{tabular}}
\end{table*}

\subsubsection{Experiments on Transparent Object Detection}\label{res_trans}

\parhead{Datasets and Metrics.} For transparent object detection, we utilize two benchmarks: glass object dataset~\textit{GDD}~\cite{mei2020donGDD} and mirror dataset~\textit{MSD}~\cite{yang2019myMSD}. The training protocol follows the setup of EBLNet~\cite{he2021enhancedEBLNet}. Quantitative evaluation is conducted using three commonly adopted metrics: mean absolute error ($M$), mean Intersection over Union (mIoU), and weighted F-measure ($F^\omega_\beta$)
Superior performance is indicated by lower values of $M$ and higher values of $F^\omega_\beta$ and $\text{mIoU}$.

\begin{table*}[!t]
    \setlength{\abovecaptionskip}{0cm}
    \setlength{\belowcaptionskip}{-0.2cm}
    \centering
    \caption{Quantitative performance gains on transparent object detection. Improvement rates (\%) are marked in \textbf{bold}.}
    \label{table:Results_GDD_MSD}
    \vspace{1mm}
    \resizebox{0.7\textwidth}{!}{
    \setlength{\tabcolsep}{2.5mm}
    \begin{tabular}{l|ccc|ccc}
        \toprule
        \multicolumn{1}{c|}{\multirow{2}{*}{Methods}} 
        & \multicolumn{3}{c|}{\textit{GDD}} 
        & \multicolumn{3}{c}{\textit{MSD}} \\
        \cline{2-7}
        & {\cellcolor{gray!40}$M$$\downarrow$} 
        & {\cellcolor{gray!40}$F^\omega_\beta$$\uparrow$} 
        & {\cellcolor{gray!40}$\text{mIoU}$$\uparrow$}
        & {\cellcolor{gray!40}$M$$\downarrow$} 
        & {\cellcolor{gray!40}$F^\omega_\beta$$\uparrow$} 
        & {\cellcolor{gray!40}$\text{mIoU}$$\uparrow$} \\
        \midrule

        EBLNet~\cite{he2021enhancedEBLNet} & 0.059 & 0.894 & 0.877 & 0.054 & 0.820 & 0.795 \\
        \rowcolor{gray!30}
        \textit{+UMBD} & 0.058 & 0.921 & 0.883 & 0.052 & 0.843 & 0.800 \\
        
        \textbf{Improve (\%)}& \textbf{1.70} & \textbf{3.02} & \textbf{0.68} & \textbf{3.70} & \textbf{2.80} & \textbf{0.63} \\
        \bottomrule
    \end{tabular}}
\end{table*}
\begin{figure}[!t]
  \centering
  \includegraphics[width=1.0 \textwidth]{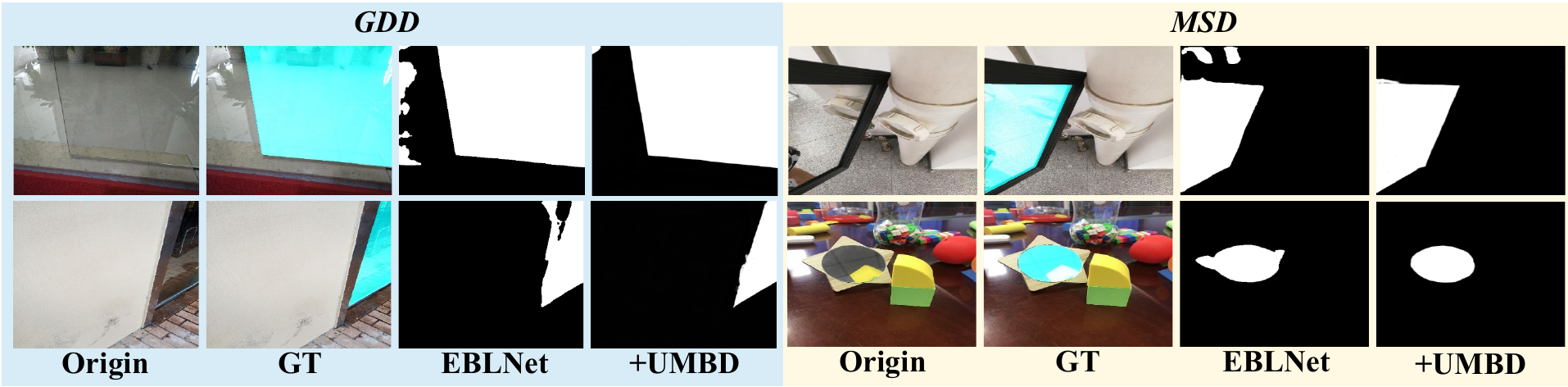}
  \vspace{-5mm}
  \caption{Qualitative refinement results on transparent object detection. Zoom in for a better view.
  }
\label{fig:methods_refinement_trans}
  \vspace{-3mm}
\end{figure}

\parhead{Results and Comparative Evaluation.} As demonstrated by the quantitative results in Tab.~\ref{table:Results_GDD_MSD} and qualitative visualization in Fig.~\ref{fig:methods_refinement_trans}, our UMBD framework achieves notable refinement gains in segmentation results from EBLNet~\cite{he2021enhancedEBLNet}. The results in Fig.~\ref{fig:methods_refinement_trans} validate the effectiveness of the proposed UMBD in transparent object detection, particularly its remarkable capability to correct substantial missegmented areas in coarse masks, further highlighting UMBD’s generalizability to broader concealed object scenarios.

\subsection{Additional Qualitative Results in COD}\label{more_res}
\begin{figure}[!h]
  \centering
  \includegraphics[width=1.0\linewidth]{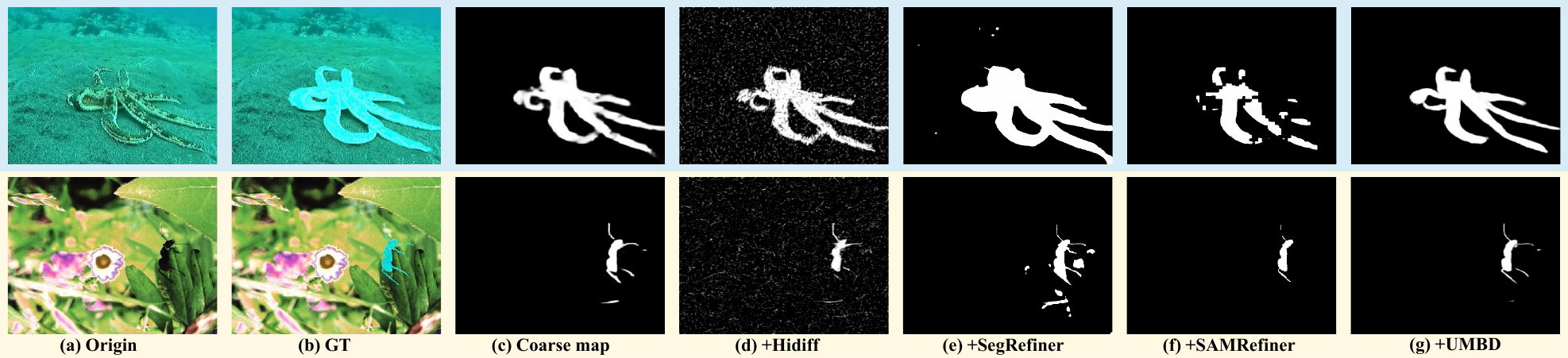}
  \vspace{-3mm}
  \caption{Qualitative comparisons with other refiners on COD benchmarks. Blue/yellow backgrounds indicate refined outputs from SINetV2 and FEDER respectively. Please zoom in for a better view.}
\label{fig:Visual_comparison_with_different_refiner_left}
  \vspace{-3mm}
\end{figure}

\subsection{Model Details}
\parhead{Details of ResBlock Transformation.}\label{app:ResBlock-transformation}Given hierarchical pretrained features $F_i^f \in \mathbb{R}^{C\times H\times W}$ and the time embedding $t \in \mathbb{R}^{d_t}$, the residual block $\mathcal{RB}$ in Eq.~\ref{eq:resblocks} enhances temporal features, yielding embedded features $\in \mathbb{R}^{C'\times H\times W}$ via Eq.~\ref{eq:detailed resblock}, where $C'$ is empirically set to 256.
\begin{equation}\label{eq:detailed resblock}
\begin{aligned}
\mathcal{RB}(F^f_i,t) &= \psi_{3\times3}^1 \circ \sigma_{\text{SiLU}} \circ \text{Norm} \Big( \text{Norm}(\psi_{3\times3}^1(F^f_i)) \odot (1+\gamma) + \beta \Big) \\
&\quad + \underbrace{\begin{cases} 
F^f_i, & C'=C \\
\psi_{1\times1}^1(F^f_i), & C'\neq C
\end{cases}}_{\text{Skip connection}}
\end{aligned}
\end{equation}
where time-conditioned parameters $\gamma,\beta \in \mathbb{R}^{C'}$ are derived from $\text{Linear}(\sigma_{\text{SiLU}}(t)) \in \mathbb{R}^{2C'} \xrightarrow{\text{split}} (\gamma, \beta)$ (shown in Eq.~\ref{eq:split}), $\sigma_{\text{SiLU}}$ denotes Sigmoid Linear Unit activation, $\text{Norm}(\cdot)$ indicates channel-wise normalization, and $\psi_{k\times k}^s$ represents convolution with kernel size $k$ and stride $s$. The symbol $\odot$ denotes element-wise multiplication.
\begin{equation}\label{eq:split}
\gamma(t), \beta(t) = \Gamma(\text{MLP}(t)), \quad \Gamma: \mathbb{R}^{2C'} \to \mathbb{R}^{C'} \times \mathbb{R}^{C'}
\end{equation}

\parhead{Details of Convolutional Fusion Module.}\label{app:stacked_CF}
The Convolutional Fusion module (CF) generates an attention map $M_{\text{attn}}$ based on Eq.~\eqref{eq: attention}:
\begin{equation}
    M_{\text{attn}}= \text{Sigmoid}\left(\psi_{1\times1}^1\left(\relu\left(\psi_{3\times3}^1\left(\psi_\text{stack}\left(\psi_\text{stack}\left(M_{\text{cat}}\right)\right)\right)\right)\right)\right),
    \label{eq: attention}
\end{equation}
where the repeated conv-block is defined as $\psi_\text{stack} = \relu(\psi_{3\times3}^2(\relu(\psi_{3\times3}^1(\cdot))))$. 

\parhead{Model Architecture of Denoiser.}\label{app:Model-Architecture}
Following~\cite{wang2023segrefiner}, we employ U-Net for our denoising network. We modify the U-Net to take in 5-channel input (concatenation of input image $\img{x}$, prior coarse mask $M_c$, and latent variable $\img{y}_t$) and output a 1-channel estimated noise $\hat{\epsilon}$. All others remain unchanged other than the aforementioned modifications.

\subsection{Related Works on COD}\label{app:codworks}
Early advancements in COD primarily relied on CNN-based architectures. For instance, SINet~\cite{fan2020CamouflageSINet} establishes a robust baseline through a simple yet effective search-identification framework, leveraging hierarchical feature extraction to tackle the intrinsic similarity between camouflaged objects and backgrounds. Building on CNNs,  FEDER~\cite{he2023camouflaged_feder} introduces learnable wavelet-based feature decomposition and edge reconstruction modules, explicitly addressing ambiguous boundaries via frequency attention and auxiliary edge-guided optimization. While CNNs excel in local feature modeling, their limited global context awareness motivates transformer-based approaches. FSPNet~\cite{Huang2023Feature} enhances locality modeling via non-local token interactions and aggregates multi-scale features through a shrinkage pyramid decoder, effectively capturing subtle cues from indistinguishable backgrounds. Recently, generative models such as diffusion models have emerged as promising alternatives for COD. FocusDiffuser~\cite{zhao2024focusdiffuser} pioneers this direction by integrating a Boundary-Driven LookUp module and Cyclic Positioning mechanism into a denoising framework, enabling iterative refinement of camouflaged object details through generative priors. These works collectively highlight the evolution from discriminative to generative paradigms in COD research.

\subsection{Broader impacts}\label{app:Broaderimpacts}
Camouflaged object detection and refinement represent fundamental challenges in computer vision, addressing the segmentation of targets intentionally concealed within their environments. In practical applications, robust COD systems significantly enhance visual perception capabilities for biodiversity monitoring~\cite{wang2024depth_agricultural}, medical anomaly detection~\cite{fan2020pranetpolyp_segmentation}, and industrial quality control~\cite{kumar2008computerdefect_detection}. Our proposed UMBD framework demonstrates particular value because precise segmentation of camouflaged objects can directly impact operational effectiveness.The inherent plug-and-play structure underscores the effectiveness and adaptability of the proposed refinement framework not only for the COD task but also potentially for other vision segmentation tasks. Notably, COD tasks and segmentation refinement works have yet to exhibit negative social impacts.
 Our proposed refinement methods also do not present any foreseeable negative societal
 consequences.

\end{document}